\makeatletter\def\graphicscache@inhibit{true}\makeatother

\documentclass[letterpaper, 10 pt, conference]{ieeeconf}  %

\IEEEoverridecommandlockouts                              %

\usepackage[utf8]{inputenc}
\usepackage[T1]{fontenc}
\usepackage[hidelinks]{hyperref}
\usepackage{graphicx}
\usepackage{tabularx}
\usepackage{multirow}
\usepackage{amsmath,amssymb}
\usepackage{xspace}
\usepackage{balance}
\usepackage[per-mode=symbol,binary-units=true,range-units=single,range-phrase=-,detect-weight=true,detect-family=true]{siunitx}
\usepackage{booktabs}
\usepackage{paralist}
\usepackage{subcaption}
\usepackage[font=footnotesize]{caption}
\usepackage{url}
\usepackage{tikz,tkz-euclide}
\usepackage{pgfplots}
\pgfplotsset{compat=1.15}
\usetikzlibrary{fit,shapes.callouts,shapes.geometric,backgrounds,positioning,arrows.meta, calc}
\usepackage[ruled,vlined]{algorithm2e}

\usepackage{graphicscache}

\newcommand{\ie}{i.e.,\ }
\newcommand{\eg}{e.g.,\ }
\newcommand{\cf}{cf.\xspace}
\newcommand{\etal}{\xspace{}et al.\xspace}

\newcommand{\reffig}[1]{Fig.~\ref{#1}}
\newcommand{\reftab}[1]{Tab.~\ref{#1}}
\newcommand{\refsec}[1]{Sec.~\ref{#1}}
\DeclareMathOperator{\atantwo}{atan2}
\DeclareMathOperator{\acos}{acos}

\title{\LARGE \bf
Predictive Angular Potential Field-based Obstacle Avoidance for Dynamic UAV Flights
}

\author{Daniel Schleich and Sven Behnke%
\thanks{All authors are with the Autonomous Intelligent Systems group, %
		University of Bonn, Germany;
        {\tt\small schleich@ais.uni-bonn.de}}%
\thanks{ This work has been funded by the German Federal Ministry of Education and Research (BMBF) in the project ``Kompetenzzentrum: Aufbau des Deutschen Rettungsrobotik-Zentrums (A-DRZ)''.}%
}

\begin{document}

\maketitle
\thispagestyle{empty}
\pagestyle{empty}

\begin{tikzpicture}[remember picture,overlay]
\node[anchor=north west,align=left,font=\sffamily,xshift=0.5cm,yshift=-0.5cm] at (current page.north west) {%
\footnotesize \textbf{Accepted final version.} IEEE/RSJ International Conference on Intelligent Robots and Systems (IROS), Kyoto, Japan, to appear October 2022. };
\end{tikzpicture}%
\begin{abstract}
In recent years, unmanned aerial vehicles (UAVs) are used for numerous inspection and video capture tasks.
Manually controlling UAVs in the vicinity of obstacles is challenging, however, and poses a high risk of collisions.
Even for autonomous flight, global navigation planning might be too slow to react to newly perceived obstacles. Disturbances such as wind might lead to deviations from the planned trajectories.

In this work, we present a fast predictive obstacle avoidance method that does not depend on higher-level localization or mapping and maintains the dynamic flight capabilities of UAVs.
It directly operates on LiDAR range images in real time and adjusts the current flight direction by computing angular potential fields within the range image.
The velocity magnitude is subsequently determined based on a trajectory prediction and time-to-contact estimation.

Our method is evaluated using Hardware-in-the-Loop simulations.
It keeps the UAV at a safe distance to obstacles, while allowing higher flight velocities than previous reactive obstacle avoidance methods that directly operate on sensor data.
\end{abstract}

\section{Introduction}
\label{sec:introduction}
Unmanned aerial vehicles (UAVs) are increasingly applied to many different tasks requiring observation of objects or environments that are difficult to access.
This includes fields like industrial inspection, agriculture, and search and rescue.
Their ability for agile flight and high velocities makes UAVs especially suitable for time-critical applications like target tracking and disaster response.
Such scenarios often require flights close to obstacles or even indoors, thus posing significant strain on human pilots during manual operation.
The application of autonomous flight systems is also challenging, however, since one has to deal with large, initially unknown environments.
In our previous work \cite{schleich2021icuas}, we presented a hierarchical navigation and control pipeline for autonomous flights in GNSS-denied environments.
We now propose an additional reactive obstacle avoidance layer (\reffig{fig:system_overview}) that can be added to our system to increase flight safety in the vicinity of obstacles.

\begin{figure}
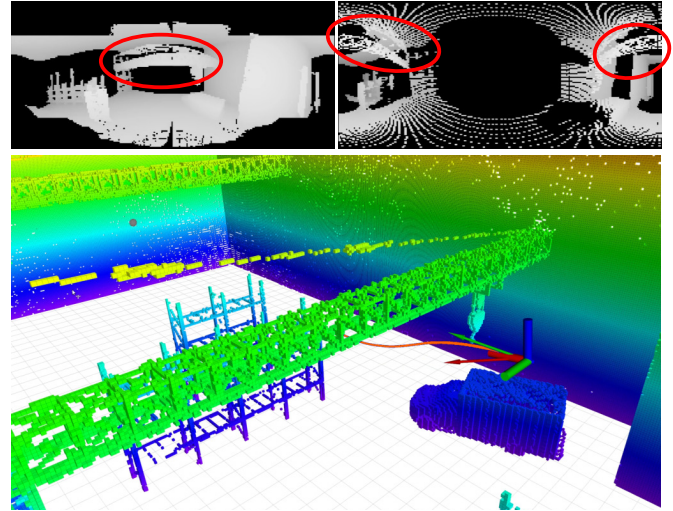

\begin{center}
\begin{tikzpicture}[
 	font=\sffamily\footnotesize,
    every node/.append style={text depth=.2ex},
	l/.style={font=\sffamily\scriptsize},
]

\def\imageWidth{4.3};
\def\imageHeight{4.8};

\node[anchor=south west,inner sep=0] (image_range) at (0,\imageHeight) {\includegraphics[width=\imageWidth cm, clip, trim=0 0 0 0]{figures/teaser/teaser_range.png}};
\node[anchor=south west,inner sep=0] (image_pred) at (4.35cm,\imageHeight) {\includegraphics[width=\imageWidth  cm, clip, trim=0 0 0 0]{figures/teaser/teaser_prediction.png}};
\node[anchor=south west,inner sep=0] (image_3d) at (0,0) {\includegraphics[width=\linewidth, clip, trim=0 50 0 0]{figures/teaser/teaser_3d.png}};

\draw[line width=0.5mm, red] (2, \imageHeight + 1.2) circle (0.8cm and 0.35cm);
\draw[line width=0.5mm, red,rotate=-10] (3.8, \imageHeight + 2.2 ) circle (0.75cm and 0.35cm);
\draw[line width=0.5mm, red,rotate=10] (9.2, \imageHeight -0.2 ) circle (0.5cm and 0.35cm);

\end{tikzpicture}
 \end{center}
\caption{ Our predictive obstacle avoidance method successfully avoids a collision. The UAV (coordinate axes) receives a velocity command (green arrow) that would result in a collision with an obstacle. Our method adjusts the flight direction using angular potential fields, which are computed on aggregated LiDAR range images (top left). Note that distant obstacles are pruned since they do not pose a risk of collision. The velocity magnitude of the final command (red arrow) is chosen based on a time-to-contact estimate along the future trajectory (orange), which is generated by iteratively applying potential fields to range images transformed into the future sensor frames.
The predicted range image for the end of the unrolled trajectory is shown on the top right. Note that the obstacle (circled red) is now behind the UAV.}
\vspace{-0.5cm}
\label{fig:teaser}
\end{figure}

Obstacle avoidance methods must be able to react quickly to unknown obstacles without depending on other modules like localization or mapping, thus ensuring safe flights even if external disturbances or errors in higher-level modules occur.
Our method can be directly applied to LiDAR range images without requiring a local map.
However, we still aggregate LiDAR scans over a short time interval into a history range image.
This is necessary to be able to avoid small structures like cables, which might not be measured in every scan.
Since we only aggregate over a short time interval, the transformation between different LiDAR range images can be approximated by integrating velocities.
Thus, we do not depend on global position estimates but only assume that an estimate of the current UAV velocity is available, which is also needed for the low-level controller to track the velocity commands generated by our method.

A key advantage for the application of UAVs is their ability for high flight velocities and thus short mission execution times.
Many obstacle avoidance methods unnecessarily limit the flight velocity, though.
For example, the repulsive forces of classic potential field methods do not only adjust the flight direction but also decelerate the UAV, although obstacles might be passed at a safe distance.
To address this issue, we propose a novel method that adjusts the flight direction based on an angular potential field directly defined on LiDAR range images.
To choose an appropriate velocity magnitude, the future trajectory is subsequently predicted to estimate the time-to-contact and the velocity command is scaled accordingly.

In summary, our proposed method includes:
\begin{itemize}
 \item the aggregation of LiDAR range images over a short time horizon into a history range image, which does not depend on global localization or mapping modules,
 \item an angular potential field method with dynamic consideration that is applied to range images to determine the flight direction, and
 \item trajectory prediction with time-to-contact estimation to scale the velocity magnitude.
\end{itemize}
 
\section{Related Work}
\label{sec:related_work}
Due to their long planning times, global trajectory planning methods cannot be executed at the high frequencies needed to avoid dynamic or previously unknown obstacles during fast UAV flight.
Thus, many approaches to low-level obstacle avoidance locally adjust a global trajectory to newly perceived obstacles.
For example, Oleynikova~\etal\cite{oleynikova2016iros} fit continuous-time polynomials to a global geometric path, which are locally optimized with respect to obstacles and control costs.
A similar method was proposed by Usenko~\etal\cite{usenko2017iros} using B-Splines instead.

Zhang~\etal\cite{zhang2018iros} use a hierarchy of multiple precomputed offline trajectories to quickly react to previously unknown obstacles in cluttered environments. 
If a collision for the currently tracked path is detected, they efficiently switch to an alternative trajectory that locally avoids obstacles.
This method significantly reduces onboard computations but depends on a prior map of the environment.
Thus, it is extended in \cite{zhang2019iros} to model parts of the environment probabilistically if they are not currently covered by onboard sensors.
Instead of searching for the shortest path, the trajectory with highest probability of reaching the goal is executed.
For the case where no prior map is known, a heuristic is introduced to estimate this probability.

All of the above methods aggregate an environment map and thus depend on global position estimates.
To ensure collision-free flights even if errors in higher-level localization modules occur, obstacle avoidance methods should directly operate on sensor data.
For example, Beul and Behnke~\cite{beul2020ssrr} generate time-optimal trajectories which are checked for collisions against LiDAR point clouds.
If necessary, additional trajectories to alternative waypoints are computed until a collision-free trajectory is found.
However, the target points do not depend on the environment structure but are sampled around the initial trajectory and the current UAV position.
Thus, the performance depends on the parametrization of the sampling process and it is prone to local minima in constrained environments.

A common method for low-level obstacle avoidance is the Dynamic Window Approach (DWA) introduced by Fox~\etal\cite{fox1997ram}.
A set of control inputs is sampled and each is evaluated by predicting the corresponding future trajectory.
The best control input is chosen based on obstacle clearance, progress towards the goal, and velocity.
Multiple extensions to DWA have been introduced.
For example, Missura and Bennewitz~\cite{missura2019icra} propose a dynamic collision model that predicts the motion of obstacles and thus can handle non-static environments.
Dobrevski~\etal\cite{dobrevski2020iros} adjust DWA parameters dynamically based on the current environment perception using a neural network and reinforcement learning.
Due to sampling the control inputs and a limited planning horizon, DWA-based methods are prone to local minima.
To increase the look-ahead, Missura~\etal\cite{missura2022iros} propose short-term aborting A* searches, but they again depend on global localization and environment mapping.

Instead of searching for a collision-free trajectory, artificial potential field-based methods \cite{khatib1986ijrr} are commonly used to instantaneously adjust movement commands to the most recent sensor measurements.
Nieuwenhuisen~\etal\cite{nieuwenhuisen2013predictive} combine potential field-based obstacle avoidance with a learned motion model to predict the future trajectory and reduce the current velocity if necessary.
In \cite{beul2018ral}, the requirement to learn a motion model is removed by introducing two different influence spheres around the robot: A larger, passive avoidance sphere, where the UAV is decelerated in the obstacle direction, and an inner sphere, where the UAV is actively pushed away from the obstacle.
Thus, smoother trajectories through narrow corridors are achieved.
However, classic potential field-based methods do not scale well to high velocities and are prone to local minima.

The idea to choose the movement direction based on range images has been proposed before, \eg by Sezer~\etal\cite{sezer2012ras}, who steer a vehicle towards the obstacle gap with largest angle in the current scan.
Houshyari and Sezer~\cite{houshyari2021robotica} extend this approach to use the Euclidean distance instead of angles to measure the gap width.
Cho~\etal\cite{cho2018jat} define a Gaussian potential field on range images.
In contrast to our approach, these methods have only been applied to 2D planning for ground vehicles and they do not consider the vehicle's dynamics.

We propose a reactive obstacle-avoidance method that directly operates on 3D LiDAR range images and does not depend on global localization or mapping.
We use velocity-dependent angular potential fields to determine the flight direction and scale the velocity magnitude by predicting the future trajectory and a time-to-contact estimation.

\section{Method}
\label{sec:method}
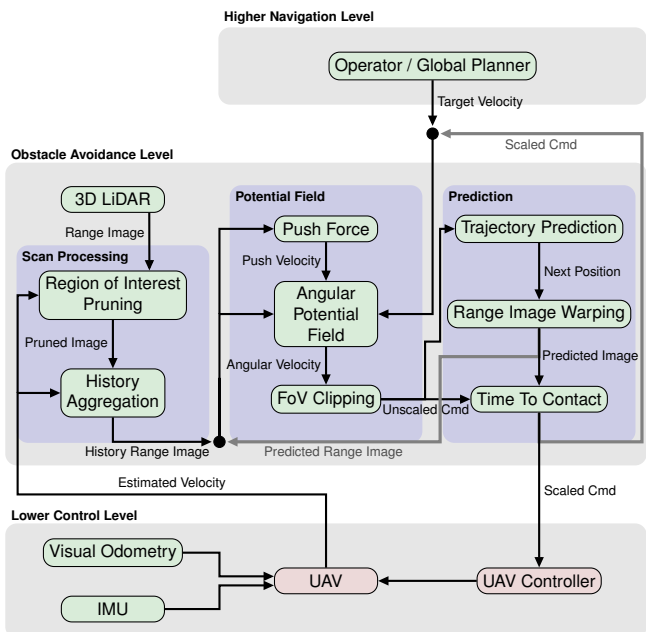
\begin{figure}
\begin{tikzpicture}
[content_node/.append style={font=\sffamily,minimum size=1.5em,minimum width=6em,draw,align=center,rounded corners,scale=0.65},
label_node/.append style={font=\sffamily,scale=0.5},
group_node/.append style={font=\sffamily,dotted,align=center,rounded corners,inner sep=1em,thick},>={Stealth[inset=0pt,length=4pt,angle'=45]}]

\definecolor{red}{rgb}     {0.5,0.0,0.0}
\definecolor{green}{rgb}   {0.0,0.5,0.0}
\definecolor{blue}{rgb}    {0.0,0.0,0.5}
\definecolor{grey}{rgb}    {0.5,0.5,0.5}

\def\width{8.5};
\def\lowHeight{1.5};
\def\obstacleHeight{4};
\def\highHeight{1};
\def\marginHeight{0.8};
\def\marginWidth{0.3};
\def\subMarginHeight{0.3};
\coordinate (lower_level) at (0, 0);
\coordinate (obstacle_avoidance_level) at (0, \lowHeight+\marginHeight);
\coordinate (higher_level) at (1/3 * \width, \lowHeight+\obstacleHeight+2*\marginHeight);

\draw[thick, rounded corners, grey!20!white,fill] (higher_level) -- ++(2/3*\width,0) -- ++(0,\highHeight) -- ++(-2/3*\width,0) -- cycle;
\draw[thick, rounded corners, grey!20!white,fill] (obstacle_avoidance_level) -- ++(\width,0) -- ++(0,\obstacleHeight) -- ++(-\width,0) -- cycle;
\draw[thick, rounded corners, grey!20!white,fill] (lower_level) -- ++(\width,0) -- ++(0,\lowHeight) -- ++(-\width,0) -- cycle;

\node(Higher_Group_Label)[label_node,anchor=south west] at (1/3 * \width,\lowHeight+\obstacleHeight+\highHeight+2*\marginHeight) {\textbf{Higher Navigation Level}};
\node(Avoidance_Group_Label)[label_node,anchor=south west] at (0,\lowHeight+\obstacleHeight+\marginHeight) {\textbf{Obstacle Avoidance Level}};
\node(UAV_Group_Label)[label_node,anchor=south west] at (0,\lowHeight) {\textbf{Lower Control Level}};

\node(Operator)[content_node,fill=green!15!white] at (2/3*\width,\lowHeight+\obstacleHeight+2*\marginHeight+0.5*\highHeight) {Operator / Global Planner};

\def\subHeight{(\obstacleHeight-2*\subMarginHeight)};
\def\subY{(\lowHeight+\obstacleHeight+\marginHeight - \subMarginHeight)};
\def\subHeightS{(0.8*\obstacleHeight-2*\subMarginHeight)};
\def\subYS{(\lowHeight+0.8*\obstacleHeight+\marginHeight - \subMarginHeight)};

\node(Lidar)[anchor=north, content_node,fill=green!15!white] at (1/6*\width,{\subY}) {3D LiDAR};

\coordinate (Scan) at (1/2 * \marginWidth,{\subYS});
\coordinate (PF) at (1/3 * \width + 1/2 * \marginWidth,{\subY});
\coordinate (Prediction) at (2/3* \width + 1/2 * \marginWidth,{\subY});

\draw[thick, rounded corners, blue!20!white,fill] (Scan) -- ++(1/3* \width -\marginWidth,0) -- ++(0,{-\subHeightS}) -- ++({-(1/3 * \width - \marginWidth)},0) -- cycle;
\draw[thick, rounded corners, blue!20!white,fill] (PF) -- ++(1/3 * \width - \marginWidth,0) -- ++(0,{-\subHeight}) -- ++({-(1/3 * \width - \marginWidth)},0) -- cycle;
\draw[thick, rounded corners, blue!20!white,fill] (Prediction) -- ++(1/3 * \width -  \marginWidth,0) -- ++(0,{-\subHeight}) -- ++({-(1/3 * \width - \marginWidth)},0) -- cycle;

\node(Scan_Group_Label)[label_node,anchor=north west] at (1/2 * \marginWidth,{\subYS}) {\textbf{Scan Processing}};
\node(PF_Group_Label)[label_node,anchor=north west] at (1/3 * \width + 1/2 * \marginWidth,{\subY}) {\textbf{Potential Field}};
\node(Prediction_Group_Label)[label_node,anchor=north west] at (2/3 * \width + 1/2 * \marginWidth,{\subY}) {\textbf{Prediction}};

\node(ScanPruning)[content_node,fill=green!15!white] at (1/6 * \width,{\subYS - 1/4 * \subHeightS}) {Region of Interest\\Pruning};
\node(History)[content_node,fill=green!15!white] at (1/6 * \width,{\subYS - 3/4 * \subHeightS}) {History\\Aggregation};

\node(Push)[content_node,fill=green!15!white] at (3/6 * \width,{\subY - 1/6 * \subHeight}) {Push Force};
\node(AnglePF)[content_node,fill=green!15!white] at (3/6 * \width,{\subY - 3/6 * \subHeight}) {Angular\\Potential\\Field};
\node(FoV)[content_node,fill=green!15!white] at (3/6 * \width,{\subY - 5/6 * \subHeight}) {FoV Clipping};

\node(Trajectory)[content_node,fill=green!15!white] at (5/6 * \width,{\subY - 1/6 * \subHeight}) {Trajectory Prediction};
\node(Warping)[content_node,fill=green!15!white] at (5/6 * \width,{\subY - 3/6 * \subHeight}) {Range Image Warping};
\node(TimeToContact)[content_node,fill=green!15!white] at (5/6 * \width,{\subY - 5/6 * \subHeight}) {Time To Contact};

\coordinate(ImageCross) at (1/3*\width,{\subY -  \subHeight});
\draw[fill=black] (ImageCross) circle (0.075);

\coordinate(TargetCross) at (2/3*\width,{\lowHeight+\obstacleHeight+1.5*\marginHeight});
\draw[fill=black] (TargetCross) circle (0.075);

\node(IMU)[content_node,fill=green!15!white] at (1/6 * \width, 1/4 * \lowHeight) {IMU};
\node(VO)[content_node,fill=green!15!white] at (1/6 * \width, 3/4 * \lowHeight) {Visual Odometry};
\node(UAV)[content_node,fill=red!15!white] at (3/6 * \width, 1/2 * \lowHeight) {UAV};
\node(UAV_Controller)[content_node,fill=red!15!white] at (5/6 * \width, 1/2 * \lowHeight) {UAV Controller};

\draw[->, thick] (Lidar.340) -- (ScanPruning.34) node[label_node, midway, left, shift={(0,0.2)}] {Range Image};
\draw[->, thick] (ScanPruning) -- (History) node[label_node, midway, left] {Pruned Image};
\draw[->, thick] (History) -- (1/6 * \width,{\subY - \subHeight}) -- (1/3*\width-0.1,{\subY -  \subHeight}) node[label_node, midway, below, shift={(-0.4,0.0)}] {History Range Image};
\draw[->, thick] (ImageCross) -- (1/3*\width,{\subY - 3/4 * \subHeight}) -- (1/3*\width,{\subY -  \subHeight})-- (1/3*\width,{\subY - 3/6 * \subHeight}) -- (AnglePF);

\draw[->, thick] (1/3*\width,{\subY - 3/6 * \subHeight}) -- (1/3*\width,{\subY - 1/6 * \subHeight}) -- (Push);

\draw[->, thick] (Push) -- (AnglePF) node[label_node, midway, left] {Push Velocity} ;
\draw[->, thick] (Operator) -- (2/3*\width,{\lowHeight+\obstacleHeight+1.5*\marginHeight+0.1}) node[label_node, midway, right] {Target Velocity};
\draw[->, thick] (TargetCross) -- (2/3 * \width,{\subY - 3/6 * \subHeight}) -- (AnglePF);
\draw[->, thick] (AnglePF) -- (FoV) node[label_node, midway, left] {Angular Velocity};
\draw[->, thick] (FoV) -- (4/6 * \width -0.1,{\subY - 5/6 * \subHeight}) -- (4/6 * \width -0.1,{\subY - 4/6 * \subHeight + 0.2}) -- (4/6 * \width +0.1,{\subY - 4/6 * \subHeight + 0.2}) -- (4/6 * \width +0.1,{\subY - 1/6 * \subHeight}) -- (Trajectory);
\draw[->, thick] (FoV) -- (TimeToContact) node[label_node, midway, below] {Unscaled Cmd};
\draw[->, thick] (Trajectory) -- (Warping) node[label_node, midway, right] {Next Position};

\draw[->, very thick, black!50] (Warping) -- (5/6 * \width,{\subY - 4/6 * \subHeight}) -- (4/6 * \width + 0.1,{\subY - 4/6 * \subHeight}) -- (4/6 * \width + 0.1,{\subY -  \subHeight}) -- (1/3*\width+0.1,{\subY -  \subHeight}) node[label_node, midway, below] {\textcolor{black!70}{Predicted Range Image}};
\draw[->, thick] (Warping) -- (TimeToContact) node[label_node, midway, right, shift={(-0.05,0.)}] {Predicted Image};

\draw[->, very thick, black!50] (TimeToContact) -- (5/6 * \width,{\subY -  \subHeight}) -- (6/6 * \width - 0.05,{\subY -  \subHeight}) -- (6/6 * \width - 0.05,{\subY -  \subHeight})  -- (\width-0.05,{\lowHeight+\obstacleHeight+1.5*\marginHeight}) -- (2/3*\width+0.1,{\lowHeight+\obstacleHeight+1.5*\marginHeight}) node[label_node, midway, below] {\textcolor{black!70}{Scaled Cmd}};
\draw[->, thick] (TimeToContact) -- (UAV_Controller) node[label_node, midway, right] {Scaled Cmd};

\draw[->, thick] (VO) -- (2/6 * \width, 3/4 * \lowHeight) --  (2/6 * \width,-1.0 |- UAV.175)  -- (UAV.175);
\draw[->, thick] (IMU)  -- (2/6 * \width, 1/4 * \lowHeight) -- (2/6 * \width,-1.0 |- UAV.185) -- (UAV.185);
\draw[->, thick] (UAV_Controller) -- (UAV);
\draw[->, thick] (UAV) -- (1/2 * \width, \lowHeight+ 1/2 * \marginHeight) -- (0.15, \lowHeight+ 1/2 * \marginHeight) node[label_node, midway, above, shift={(0.0,0.)}] {Estimated Velocity} -- (0.15, {\subYS - 3/4 * \subHeightS}) -- (History);
\draw[->, thick] (0.15, {\subYS - 3/4 * \subHeightS}) -- (0.15, {\subYS - 1/4 * \subHeightS}) -- (ScanPruning);

\end{tikzpicture}
  \caption{System overview. The computation blocks \textit{Potential Field} and \textit{Prediction} are executed iteratively to unroll the future trajectory. Black dots mark the entry points to these iterations. Starting from the second iteration, the results of the previous iteration (gray arrows) are forwarded instead of the initial input.}
 \label{fig:system_overview}
 \vspace{-0.2cm}
\end{figure}

In the following sections, we give a detailed description of the different components of our obstacle avoidance approach depicted in \reffig{fig:system_overview}.
First, the current 3D LiDAR scan is preprocessed (\refsec{sec:scan_processing}) by reducing the resolution and removing distant points that do not pose a risk of a collision in the near future.
Additionally, we aggregate the current range image with the previous ones over a short time horizon to be able to avoid small obstacles that are not measured in every scan.

In order to keep safe distances to obstacles while still allowing fast flights on collision-free trajectories in narrow passages, we represent velocity commands in spherical coordinates and split their generation into two parts:
In \refsec{sec:angular_potential_field}, we determine the flight direction, \ie the angular components, by applying a potential field method to the sensor range image.
Subsequently, we predict the future trajectory (\refsec{sec:prediction}) by iteratively applying our potential field method and unrolling the resulting velocity commands (scaled with the commanded target velocity).
Thus, we can detect future collisions and scale the actually executed velocity command such that the UAV will not enter the safety region around obstacles within the minimal time-to-contact $t_\text{contact}$.

\subsection{LiDAR Scan History Aggregation}
\label{sec:scan_processing}

\begin{figure}
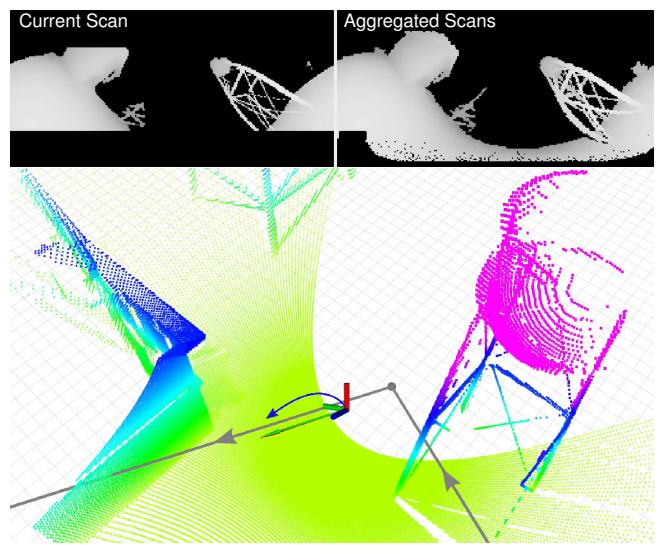

\begin{center}
\begin{tikzpicture}[
 	font=\sffamily\footnotesize,
    every node/.append style={text depth=.2ex},
	l/.style={font=\sffamily\scriptsize},
]

\def\imageHeight{5.0};
\def\RangeImageHeight{0.43955 * \imageHeight };

\node[anchor=south west,inner sep=0] (image_range) at (0,\imageHeight) {\includegraphics[width=0.99\linewidth, clip, trim=0 0 0 0]{figures/aggregation/range_images_new.png}};
\node[anchor=south west,inner sep=0] (image_3d) at (0,0) {\includegraphics[width=0.99\linewidth, clip, trim=0 0 0 0]{figures/aggregation/3d_new.png}};

\node(scan)[anchor=north west, white, font=\sffamily\scriptsize] at (0, \imageHeight + \RangeImageHeight) {Current Scan};
\node(scan)[anchor=north west, white, font=\sffamily\scriptsize] at (0.5 * \linewidth, \imageHeight + \RangeImageHeight) {Aggregated Scans};

\end{tikzpicture}
 \end{center}
\caption{ Comparison of a single range image (top left) against the aggregated range image (top right) during an aggressive change in flight direction. The corresponding 3D view of the environment is shown at the bottom including the UAV (coordinate axes), the target trajectory (gray),
the target and adjusted velocity commands (green and red arrows) as well as the predicted trajectory (blue). Due to the UAV orientation, the ground in front of the UAV is not visible in the current scan. Aggregating range images over a short time interval efficiently extends the vertical field of view. As a result, the ground is still considered during obstacle avoidance.}
 \label{fig:aggregation}
\end{figure}

Our method is able to directly operate on LiDAR range images.
Only considering the most recent scan is dangerous, however, since thin structures might not be represented in every measurement.
Furthermore, during dynamic flight, obstacles might be moved out of the sensor's limited vertical field of view.
This is illustrated in \reffig{fig:aggregation}.
During frequent aggressive changes of flight direction, a velocity command might be chosen that steers the UAV closer to currently not visible obstacles, even when restring commands to the sensor's field of view.
In combination with control latencies, this might result in the UAV entering the safety region around obstacles or even in a collision.
To address this issue, we aggregate measurements over a short time interval $t_\text{history}=\SI{1}{\second}$ into a history range image $\mathcal H$.
For each pixel $(\phi,\theta)$, we additionally keep track of the age $t(\phi,\theta)$ of the corresponding measurement.

For lower computation times, we discard all pixels of the current image with range values that are too high to pose a risk of collision in the near future.
To this end, we predict the area $A_\text{pos}$ of possible future UAV positions by applying a maximum acceleration command $\pm a_{\text{max}}$ to the current UAV velocity $v_0$ for a time interval $t=t_\text{history}+t_\text{contact}$.
Here, $t_\text{contact}$ denotes the minimum allowed time-to-contact, which equals the prediction horizon used in \refsec{sec:prediction}.
All pixels corresponding to 3D points whose distance to $A_\text{pos}$ exceed a safety margin $d_\text{safe}$ can be skipped during the processing described below.

Every time a new 3D scan is obtained, we transform the history range image $\mathcal H_{i-1}$ of the previous iteration into the new sensor frame.
First, we project every pixel $p_\text{Image}=(\phi,\theta)$ into 3D space.
For simplicity of notation, we state the transformations for continuous angles and discard the discretization into image coordinates:
\begin{equation}
\begin{split}
\mathcal T_{\text{Image} \mapsto \text{3D} }(p_\text{Image}) =& \frac{r(\phi,\theta)}{\|\overline p_\text{3D}\|} \overline p_\text{3D}, \\
\text{ with } \overline p_\text{3D} =& \begin{pmatrix} \cos(\theta)\cos(\phi) \\ \cos(\theta)\sin(\phi) \\ \sin(\theta) \end{pmatrix}.
\end{split}
 \label{eq:image_to_3d}
\end{equation}
Here, $r(\phi,\theta)$ denotes the range value associated with the pixel $(\phi,\theta)$.
The offset between the previous and current sensor frame can be estimated by integrating UAV velocity measurements.
The projected point $p_\text{3D}=(x,y,z)$ is then transformed accordingly and reprojected into the new image frame using
\begin{equation}
\mathcal T_{\text{3D} \mapsto \text{Image}}( p_\text{3D} ) = \begin{pmatrix}\atantwo(y,x) \\ \frac{\pi}{2}-\acos(\frac{z}{\|p_\text{3D}\|_2}) \end{pmatrix}.
 \label{eq:3d_to_image}
\end{equation}
Subsequently, we merge the transformed history image $\widehat{\mathcal H}_{i-1}$ with the current scan $\mathcal S_i$.
Range values and pixel ages of the history image should be updated if the corresponding obstacle is present in the current scan, although the new range value might be larger than the history value.
Thus, instead of taking the pixel-wise minimum of $\widehat{\mathcal H}_{i-1}$ and $\mathcal S_i$, we keep the most recent range values $r_{\mathcal S_i}(\phi,\theta)$ even if they are larger than the history values $r_{\widehat{\mathcal H}_{i-1}}(\phi,\theta)$, as long as the difference is below a threshold $T$.
To model the uncertainty of history values $r_{\widehat{\mathcal H}_{i-1}(\phi,\theta)}$, this threshold should be different for each pixel and should grow monotonically with increasing pixel age $t(\phi,\theta)$.
Thus, we define the range values of the updated history image $\mathcal H_i$ as
\begin{equation}
r_{\mathcal H_i(\phi,\theta)} = 
    \begin{cases}
        r_{\widehat{\mathcal H}_{i-1}}(\phi,\theta),&\text{if } r_{\widehat{\mathcal H}_{i-1}}(\phi,\theta) e^{ \frac{t(\phi,\theta)}{\tau} } \leq r_{\mathcal S_i}(\phi,\theta)  \\
        r_{\mathcal S_i}(\phi,\theta),              &\text{otherwise,}
    \end{cases}
\end{equation}
where $\tau>0$ controls the growth rate of the threshold $T=e^{ \frac{t(\phi,\theta)}{\tau} }$.

\subsection{Potential Field Method}
\label{sec:angular_potential_field}

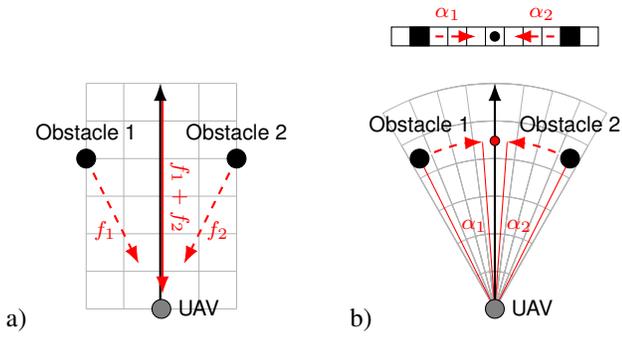
\begin{figure}
\begin{center}
a)\mbox{%
\begin{tikzpicture}[
 	font=\sffamily\footnotesize,
    every node/.append style={text depth=.2ex},
	l/.style={font=\sffamily\scriptsize},
]

\def\robotX{2};
\def\robotY{0};

\def\obsAX{1};
\def\obsAY{2};

\def\obsBX{3};
\def\obsBY{2};

\def\targetX{0};
\def\targetY{3};

\def\forceLength{0.7};
\def\radius{0.125};
\def\rangeImageHeightDist{1.5};

\coordinate (robot) at (\robotX,\robotY);
\coordinate (obstacle1) at (\obsAX,\obsAY);
\coordinate (obstacle2) at (\obsBX,\obsBY);

\coordinate (range1) at (\obsAX - 3*\radius,\obsAY + \rangeImageHeightDist);
\coordinate (range2) at (\obsBX + 3*\radius,\obsBY + \rangeImageHeightDist + 2*\radius);
\draw[white] (range1) rectangle (range2);

\def\gridStep{0.5};
\foreach \i in {2,...,6}
    \draw[ultra thin, black!30!white] (\i*\gridStep, 0) -- (\i*\gridStep, 3);
\foreach \i in {0,...,6}
    \draw[ultra thin, black!30!white] (1,\i*\gridStep) -- (3,\i*\gridStep);

\coordinate (force_end1) at ({\obsAX + \forceLength * (\robotX - \obsAX)}, {\obsAY + \forceLength * (\robotY - \obsAY)});
\coordinate (force_end2) at ({\obsBX + \forceLength * (\robotX - \obsBX)}, {\obsBY + \forceLength * (\robotY - \obsBY)});
\coordinate (total_end) at ({\robotX + \forceLength * (\robotX - \obsAX) + \forceLength * (\robotX - \obsBX)}, {\robotY + \forceLength * (\robotY - \obsAY) + \forceLength * (\robotY - \obsBY)});
\coordinate (target_end) at ({\robotX + \targetX}, {\robotY + \targetY});
\coordinate (result_end) at ({\robotX + \targetX + \forceLength * (\robotX - \obsAX) + \forceLength * (\robotX - \obsBX)}, {\robotY + \targetY + \forceLength * (\robotY - \obsAY) + \forceLength * (\robotY - \obsBY)});

\draw[-Latex,thick, dashed, red] (obstacle1) -- (force_end1) node[midway, below, shift={(-0.1,0)}] {$f_1$};
\draw[-Latex,thick, dashed, red] (obstacle2) -- (force_end2) node[midway, below, shift={(0.1,0)}] {$f_2$};
\draw[-Latex,thick, black] ([shift=({-0.015,0})]robot) -- ([shift=({-0.015,0.0})]target_end);
\draw[-Latex,thick, red] ([shift=({0.015,-0.23})]target_end) -- ++({(\forceLength * (\robotX - \obsAX) + \forceLength * (\robotX - \obsBX))},{(\forceLength * (\robotY - \obsAY) + \forceLength * (\robotY - \obsBY)) +0.23}) node[midway, above, sloped] {$f_1+f_2$};

\draw[fill=black!50] (robot) circle (\radius) node[right, shift={(0.1,0.0)}] {UAV};
\draw[fill=black] (obstacle1) circle (\radius) node[above, shift={(0.0,0.1)}] {Obstacle 1};
\draw[fill=black] (obstacle2) circle (\radius) node[above, shift={(0.0,0.1)}] {Obstacle 2};
\end{tikzpicture}
}
  \hspace{1em}
b)\hspace*{-1ex}\mbox{%
\begin{tikzpicture}[
 	font=\sffamily\footnotesize,
    every node/.append style={text depth=.2ex},
	l/.style={font=\sffamily\scriptsize},
]
    
\def\robotX{2};
\def\robotY{0};

\def\obsAX{1};
\def\obsAY{2};

\def\obsBX{3};
\def\obsBY{2};

\def\targetX{0};
\def\targetY{3};

\def\forceLength{0.7};
\def\angleRes{7.429};

\def\radius{0.125};
\def\radiusA{2.236};
\def\rangeImageHeightDist{1.5};

\coordinate (robot) at (\robotX,\robotY);
\coordinate (obstacle1) at (\obsAX,\obsAY);
\coordinate (obstacle2) at (\obsBX,\obsBY);
\coordinate (range1) at (\obsAX - 3*\radius,\obsAY + \rangeImageHeightDist);
\coordinate (range2) at (\obsBX + 3*\radius,\obsBY + \rangeImageHeightDist + 2*\radius);

\coordinate (force_end1) at ({\obsAX + \forceLength * (\robotX - \obsAX)}, {\obsAY + \forceLength * (\robotY - \obsAY)});
\coordinate (force_end2) at ({\obsBX + \forceLength * (\robotX - \obsBX)}, {\obsBY + \forceLength * (\robotY - \obsBY)});
\coordinate (total_end) at ({\robotX }, {\robotY + \radiusA});
\coordinate (target_end) at ({\robotX + \targetX}, {\robotY + \targetY});
\coordinate (result_end) at ({\robotX + \targetX + \forceLength * (\robotX - \obsAX) + \forceLength * (\robotX - \obsBX)}, {\robotY + \targetY + + \forceLength * (\robotY - \obsAY) + \forceLength * (\robotY - \obsBY)});

\def\gridStep{0.5};
\foreach \r in {1,...,6}
    \foreach \i in {1,...,4}
        \draw[ultra thin, black!30!white] (robot) -- ++(0,\r*\gridStep) arc [start angle=90, delta angle = -\i*\angleRes, radius=\r*\gridStep] -- (robot);
        
\foreach \r in {1,...,6}
    \foreach \i in {1,...,4}
        \draw[ultra thin, black!30!white] (robot) -- ++(0,\r*\gridStep) arc [start angle=90, delta angle = \i*\angleRes, radius=\r*\gridStep] -- (robot);

\draw[thin,red] (robot) -- (obstacle1) arc [start angle=180-64, delta angle = -3 * \angleRes, radius=\radiusA cm] -- (robot) node[midway, left, shift={(0.1,0.0)}] {$\alpha_1$};
\draw[thick, white] (obstacle1) arc [start angle=180-64, delta angle = -3 * \angleRes, radius=\radiusA cm];
\draw[thick,dashed,red,-Latex] (obstacle1) arc [start angle=180-64, delta angle = -3 * \angleRes, radius=\radiusA cm];

\draw[thin,red] (robot) -- (obstacle2) arc [start angle=64, delta angle = 3 * \angleRes, radius=\radiusA cm] -- (robot) node[midway, right, shift={(-0.05,0.0)}] {$\alpha_2$};
\draw[thick, white] (obstacle2) arc [start angle=64, delta angle = 3 * \angleRes, radius=\radiusA cm];
\draw[thick,dashed, red,-Latex] (obstacle2) arc [start angle=64, delta angle = 3 * \angleRes, radius=\radiusA cm];
\draw[-Latex,thick, black] (robot) -- (target_end);

\draw (range1) rectangle (range2);
\foreach \i in {1,...,10}
    \draw ({\obsAX - 3*\radius + \i * 2*\radius}, {\obsAY + \rangeImageHeightDist}) --  ++(0, 2*\radius);
\draw[-Latex,thick, dashed, red] (\obsAX,\obsAY+\rangeImageHeightDist+\radius) -- ++(6*\radius,0) node[midway, above, shift={(0,0.1)}] {$\alpha_1$};
\draw[-Latex,thick, dashed, red] (\obsBX,\obsBY+\rangeImageHeightDist+\radius) -- ++(-6*\radius,0) node[midway, above, shift={(0,0.1)}] {$\alpha_2$};
\draw[fill=black] (\obsAX-\radius,\obsAY+\rangeImageHeightDist) rectangle ++(2*\radius,2*\radius);
\draw[fill=black] (\obsBX-\radius,\obsBY+\rangeImageHeightDist) rectangle ++(2*\radius,2*\radius);
\draw[fill=black] (\robotX,\obsBY+\rangeImageHeightDist+\radius) circle (0.5*\radius);

\draw[fill=black!50] (robot) circle (\radius) node[right, shift={(0.1,0.0)}] {UAV};
\draw[fill=black] (obstacle1) circle (\radius) node[above, shift={(0.0,0.2)}] {Obstacle 1};
\draw[fill=black] (obstacle2) circle (\radius) node[above, shift={(0.0,0.2)}] {Obstacle 2};
\draw[fill=red] (total_end) circle (0.5*\radius);

\end{tikzpicture}
}
  \hspace{1em}
\end{center}
\caption{Comparison of Cartesian and spherical repulsive forces. a) Top-down view in Cartesian coordinates. The sum of the repulsive forces $f_1+f_2$ cancels out the commanded velocity (black arrow). b) Top: The angular repulsive forces $\alpha_1, \alpha_2$ are computed in image coordinates on the LiDAR range image. They cancel out each other and guide the robot towards the center of the gap between the obstacles. Bottom: Top-down view of the projection into spherical coordinates.}
\vspace{-0.3cm}
\label{fig:cartesian_vs_spherical}
\end{figure}

The repulsive forces of classical potential field methods are defined in Cartesian coordinates and thus influence both, the UAV flight direction and its velocity.
For example, consider the situation depicted in \reffig{fig:cartesian_vs_spherical}\,a, where the UAV is commanded into a gap between two obstacles.
The UAV is pushed away from each obstacle by the repulsive forces $f_1$ and $f_2$, respectively.
Depending on the force magnitudes $\|f_1\|$ and $\|f_2\|$, the commanded velocity is reduced or even completely cancelled out, resulting in a local minimum.
To address this issue, we propose to formulate the problem in spherical coordinates and apply the repulsive forces only to the angular components $\phi$ and $\theta$ (\reffig{fig:cartesian_vs_spherical}\,b).
Thus, a potential field is computed for the LiDAR range image.
The corresponding angular repulsive forces $\alpha_1,\alpha_2$ steer the UAV towards the center of the gap between the obstacles without influencing the magnitude of the resulting velocity command, which will be determined in a subsequent step based on a time-to-contact estimation.
Our experiments show that this method is less prone to local minima and allows fast flights through narrow corridors, while keeping sufficient distance to the walls.

\subsubsection{Angular Potential Field Method}
In a first step, we project the commanded target velocity $v_\text{target}$ into the range image using \eqref{eq:3d_to_image}, obtaining the target pixel $(\phi_\text{target},\theta_\text{target})$.
The angular components $(\phi_\text{out},\theta_\text{out})$ of the adjusted velocity command are then computed by adding repulsive forces $\alpha_i=(\phi_i,\theta_i)$ induced by all other pixels.

For the repulsive force computation, we define the support of a pixel as the area where its repulsive forces are non-zero.
Since the UAV should keep a safety distance $d_\text{safe}$ to each obstacle, the support of a pixel $p_\text{obs}=(\phi_\text{obs},\theta_\text{obs})$ can be chosen as the projection of a 3D sphere with radius $d_\text{safe}$ and center $\mathcal T_{\text{Image} \mapsto \text{3D} }(p_\text{Image})$ into the range image.
Depending on the range value $r$, the resulting support radius is given by $d_\text{support}=\atantwo(d_\text{safe}, r)$.
The repulsive force induced by $p_\text{obs}$ acting on a pixel $p'=(\phi',\theta')$ is then defined as
\begin{equation}
\alpha = 
    \begin{cases}
        \frac{d_\text{support} - \|p'-p_\text{obs}\|}{\|p'-p_\text{obs}\|}(p'-p_\text{obs}),&\text{if } \|p'-p_\text{obs}\| \leq d_\text{support}  \\
        0,              &\text{otherwise.}
    \end{cases}
 \label{eq:repulsive_force}
\end{equation}

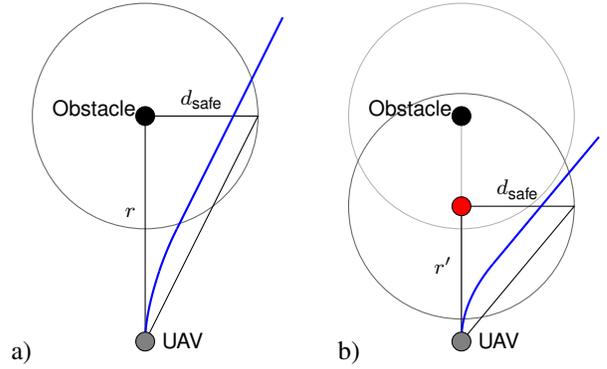
\begin{figure}
\begin{center}
a)\mbox{%
\begin
{tikzpicture}[
 	font=\sffamily\footnotesize,
    every node/.append style={text depth=.2ex},
	l/.style={font=\sffamily\scriptsize},
]
    
\def\robotX{0};
\def\robotY{0};

\def\obsX{0};
\def\obsY{3};

\def\sampleX{0};
\def\sampleY{2};

\def\radius{0.125};
\def\distSafe{1.5};

\coordinate (sample) at (\sampleX,\sampleY);
\coordinate (obstacle) at (\obsX,\obsY);
\coordinate (robot) at (\robotX,\robotY);

\draw[black!60!white] (obstacle) circle (\distSafe);

\draw (robot) -- (obstacle) node[midway, left, shift={(0.0,0.2)}] {$r$} -- ++(\distSafe,0) node[midway, above, shift={(0.0,0.0)}] {$d_\text{safe}$}-- (robot);

\def\accel{2};
\def\velAX{0};
\def\velAY{2};
\def\velBX{\distSafe};
\def\velBY{(\obsY-\robotY)};
\def\timeX{(\velBX-\velAX)/\accel};
\def\timeY{(\velBY-\velAY)/\accel};

\draw[thick, scale=0.75, domain=0:\timeY, smooth, variable=\t, blue] plot ({\robotX+\t*\velAX+0.5*\t*\t*\accel}, {\robotY+\t*\velAY+0.5*\t*\t*\accel});
\draw[thick, scale=0.75, domain=\timeY:\timeX, smooth, variable=\t, blue] plot ({\robotX+\t*\velAX+0.5*\t*\t*\accel}, {\robotY+\timeY*\velAY+0.5*\timeY*\timeY*\accel + + (\t-\timeY)*\velBY});
\draw[thick, scale=0.75, domain=\timeX:2, smooth, variable=\t, blue] plot ({\robotX+\timeX*\velAX+0.5*\timeX*\timeX*\accel + (\t-\timeX)*\velBX}, {\robotY+\timeY*\velAY+0.5*\timeY*\timeY*\accel + (\t-\timeY)*\velBY});

\draw[fill=black] (obstacle) circle (\radius) node[left, shift={(0.0,0.1)}] {Obstacle};
\draw[fill=black!50] (robot) circle (\radius) node[right, shift={(0.1,0.0)}] {UAV};

\end{tikzpicture}
}
  \hspace{1em}
b)\hspace*{-1ex}\mbox{%
\begin
{tikzpicture}[
 	font=\sffamily\footnotesize,
    every node/.append style={text depth=.2ex},
	l/.style={font=\sffamily\scriptsize},
]

\def\robotX{0};
\def\robotY{0};

\def\obsX{0};
\def\obsY{3};

\def\accel{2};
\def\velAX{0};
\def\velAY{2};

\def\sampleX{0};
\def\sampleY{\obsY-\timeToContact*\velAY};
\def\distSafe{1.5};

\def\velBX{\distSafe};
\def\velBY{(\sampleY-\robotY)};
\def\timeX{(\velBX-\velAX)/\accel};
\def\timeY{(-(\velBY-\velAY)/\accel)};
\def\timeToContact{0.6};

\def\radius{0.125};

\coordinate (robot) at (\robotX,\robotY);
\coordinate (sample) at (\sampleX,\sampleY);
\coordinate (obstacle) at (\obsX,\obsY);

\draw[black!30!white] (obstacle) circle (\distSafe);

\draw[black!30!white] (robot) -- (obstacle);

\draw[black!60!white] (sample) circle (\distSafe);
\draw (robot) -- (sample) node[midway, left, shift={(0.0,0.1)}] {$r'$} -- ++(\distSafe,0) node[midway, above, shift={(0.0,0.0)}] {$d_\text{safe}$}-- (robot);

\draw[thick, scale=0.75, domain=0:\timeY, smooth, variable=\t, blue] plot ({\robotX+\t*\velAX+0.5*\t*\t*\accel}, {\robotY+\t*\velAY+0.5*\t*\t*\accel});
\draw[thick, scale=0.75, domain=\timeY:\timeX, smooth, variable=\t, blue] plot ({\robotX+\t*\velAX+0.5*\t*\t*\accel}, {\robotY+\timeY*\velAY+0.5*\timeY*\timeY*\accel + + (\t-\timeY)*\velBY});
\draw[thick, scale=0.75, domain=\timeX:2., smooth, variable=\t, blue] plot ({\robotX+\timeX*\velAX+0.5*\timeX*\timeX*\accel + (\t-\timeX)*\velBX}, {\robotY+\timeY*\velAY+0.5*\timeY*\timeY*\accel + (\t-\timeY)*\velBY});

\draw[fill=black] (obstacle) circle (\radius) node[left, shift={(0.0,0.1)}] {Obstacle};
\draw[fill=black!50] (robot) circle (\radius) node[right, shift={(0.1,0.0)}] {UAV};
\draw[fill=red] (sample) circle (\radius);

\end{tikzpicture}
}
  \hspace{1em}
\end{center}
\caption{Definition of the support radius for repulsive force computation. Depicted is the top-down view in Cartesian coordinates of a situation where the UAV has an initial velocity towards the obstacle. a) When defining the support (gray circle) using the Euclidean range value $r$, the actual UAV trajectory (blue) passes the obstacle at a too close distance since the UAV cannot instantaneously change its flight direction. b) Using the velocity-dependent predicted future range value $r_\text{vel} := r-t_\text{contact} v$ results in larger acceleration commands. Thus, the UAV does not enter the safety area around the obstacle. }
\vspace{-0.2cm}
\label{fig:support}
\end{figure}

This definition of repulsive forces adjusts the UAV flight direction such that it has a safety distance $d_\text{safe}$ to the obstacle once the UAV reaches the obstacle depth (see \reffig{fig:support}\,a).
However, the direct line-of-sight to this position intersects with the safety area around the obstacle.
Since the UAV cannot instantaneously change its flight direction---especially at high velocities---the actual future trajectory would pass the obstacle at an even smaller distance.
To address this issue, we propose not to use the pixel range value $r$, \ie its Euclidean distance to the UAV, for the support radius definition, but a different distance metric that incorporates the UAV velocity.
Thus, we define $r_\text{vel} := r-d_\text{contact}$ as the predicted future range value after moving a distance $d_\text{contact}$ towards the obstacle.
This virtually moves the obstacle closer to the UAV, as depicted in \reffig{fig:support}\,b.
As a result, the change in flight direction will be larger for higher velocities, and a safe distance to the obstacle is maintained.
The distance $d_\text{contact}$ is defined as the predicted movement towards the obstacle $p_\text{obs}$.
Thus, it depends on the current (scalar) velocity $v$ into the obstacle direction, which is obtained using the vector projection of the current velocity vector onto the direction towards $p_\text{obs}$.
To increase the change in flight direction even if the current velocity is zero, we ensure a lower threshold $d_\text{min\_contact}$.
Thus, we define
\begin{equation}
  d_\text{contact} := \max\{t_\text{contact} v, d_\text{min\_contact}\}.
\end{equation}

For lower computation times, we additionally discard obstacles that are still far away.
The support radius of a pixel $p_\text{obs}$ with range $r$ is thus defined as
\begin{equation}
 d_\text{support} =
     \begin{cases}
        0,&\text{if } r_\text{vel} \geq d_\text{safe}  \\
        \atantwo(d_\text{safe}, r_\text{vel}),&\text{if } r_\text{vel} > 0  \\
        \frac{\pi}{2},              &\text{otherwise.}
    \end{cases}
\end{equation}

When obstacles spread over multiple pixels, the corresponding repulsive forces accumulate and adjust the flight direction stronger than necessary.
This can even lead to unstable flight trajectories with frequent large jumps in the direction command.
To avoid unbounded accumulation while still allowing repulsive forces with different directions to cancel each other out, we keep track of the minimum and maximum forces $\theta_\text{min}, \theta_\text{max},\phi_\text{min}, \phi_\text{max}$ along each dimension independently.
The sum of repulsive forces is then clipped to these values.
Thus, using the repulsive forces $\alpha_i=(\phi_i,\theta_i)$, the angular components $(\phi_\text{out},\theta_\text{out})$ of the adjusted velocity command are computed as
\begin{equation}
 \begin{pmatrix}\phi_\text{out} \\ \theta_\text{out} \end{pmatrix} = \begin{pmatrix}\phi_\text{target} \\ \theta_\text{target} \end{pmatrix} + \begin{pmatrix}\min\{\max\{\sum_i\phi_i, \phi_\text{min}\}, \phi_\text{max}\} \\ \min\{\max\{\sum_i\theta_i, \theta_\text{min}\}, \theta_\text{max}\}\end{pmatrix}.
\end{equation}

To ensure that the UAV is only moving into a direction that is covered by the LiDAR, $\theta_\text{out}$ is subsequently clipped to the vertical field-of-view.
Finally, the adjusted velocity command is transformed back into Cartesian coordinates, scaled to the target velocity magnitude.

\subsubsection{Push Force}
The angular potential field method changes the commanded flight angle at most by \SI{90}{\degree}.
If the UAV already is too close to an obstacle, adjustments of up to \SI{180}{\degree} are necessary to move the UAV away from the obstacle.
Thus, such situations have to be addressed separately.
Note that the angular potential field method was introduced to allow dynamic flights.
If the UAV is within the safety area around obstacles, we do not allow high flight velocities anymore.
Instead, the UAV should be pushed slowly away from the obstacle.
Thus, we use a classical potential field method in Cartesian coordinates for this situation.

Every time a new range images is available, we first check it for obstacles closer to the UAV than $d_\text{safe}$.
Each such obstacle induces a repulsive force with a magnitude that grows linearly with decreasing distance to the UAV.
The total push force $F_\text{push}$ is obtained as the sum of repulsive forces, scaled to the desired movement velocity.

If the UAV is closer to an obstacle than a threshold $d_\text{close} < d_\text{safe}$, the target command is completely discarded and the adjusted velocity command equals the push force.
However, if the distance to the nearest obstacle exceeds $d_\text{safe}$, the adjusted velocity command is generated by the angular potential field method described above.
To allow a smooth transition between those two cases, we combine both methods if the closest obstacle distance is between $d_\text{close}$ and $d_\text{safe}$.
Thus, we add the push force $F_\text{push}$ to the target command $v_\text{target}$.
To avoid speeding up the UAV if the directions of $F_\text{push}$ and $v_\text{target}$ align, we additionally subtract the part of the target command that steers the UAV into the direction of the push force, \ie
\begin{equation}
 v'_\text{target} = v_\text{target} + F_\text{push} - cF_\text{push},
\end{equation}
where $c$ is the scalar projection of $v_\text{target}$ onto $F_\text{push}$.
The resulting command $v'_\text{target}$ is then further adjusted by the angular potential field method described above.
To avoid oscillations where the UAV is repeatedly commanded into the safety area and pushed out again, we additionally limit the acceleration, \ie the difference between two subsequent velocity commands, if the UAV approaches the distance $d_\text{close}$.
Afterwards, the velocity command is sent to the trajectory prediction module to determine the velocity magnitude.

\subsection{Trajectory Prediction and Velocity Generation}
\label{sec:prediction}

The angular potential field method only adjusts the angular components of the flight command.
In this section, we describe how the velocity magnitude is chosen based on a time-to-contact estimation.

We scale the adjusted velocity command $v_{\text{cmd},0}$ from the potential field method to the initially commanded velocity and predict the future UAV positions.
Here, we assume that the low-level controller tries to reach the commanded velocity as fast as possible by applying a maximum acceleration $a_\text{max}$.
For a single axis with current velocity $v_0$, the time $t_\text{accel}$ needed to reach the velocity command $v_\text{cmd}$ is thus given by
\begin{equation}
 t_\text{accel} = \frac{v_\text{cmd} - v_0}{a_\text{max}}.
\end{equation}
The position after a time step $\Delta t$ is
\begin{equation}
 \begin{split}
 p' = \min\{t_\text{accel}, \Delta t\} v_0 &+ \frac{1}{2} \min\{t_\text{accel}, \Delta t\}^2 a_\text{max} \\
    &+ \max\{\Delta t - t_\text{accel}, 0\} v_\text{cmd}.
 \end{split}
\end{equation}
Note that the initial position $p_0 = 0$, since all calculations are performed in the egocentric sensor frame.
The corresponding future velocity is given by
\begin{equation}
 v' = v_0 + \min\{t_\text{accel}, \Delta t\} a_\text{max}.
\end{equation}
Using a more complex motion model would also be possible.
Obtaining an accurate model of the UAV dynamics is challenging, however, and it has to be done for every UAV independently.
Instead, we use this simplified generic model, which proved sufficiently accurate in our experiments to allow safe, dynamic flights in the vicinity of obstacles.

We do not only consider the influence of the UAV dynamics on the trajectory, but also future velocity commands generated by our potential field method.
Thus, we choose the prediction time step $\Delta t=\SI{0.05}{\second}$ corresponding to the frequency with which new LiDAR scans are obtained.
We then transform the current history range image into the predicted sensor frame as already described for the history aggregation in \refsec{sec:scan_processing} and apply the potential field method from \refsec{sec:scan_processing} to generate the next velocity command $v_{\text{cmd},1}$.
This process is iterated until we reach the prediction horizon $t_\text{contact}$ or until the closest obstacle distance is smaller than $d_\text{safe}$.
In the latter case, let $t$ denote the time of the last iteration.
Since we want the time the UAV needs to reach the safety area around obstacles to be at least $t_\text{contact}$, we scale the velocity command accordingly.
Thus, the velocity command sent to the low-level flight controller is
\begin{equation}
 v_\text{out} = \frac{t}{t_\text{contact}} v_{\text{cmd},0}.
\end{equation}
If the initial UAV position already is within the safety area, the time-to-contact based approach cannot be applied.
In this case, we execute the velocity command only if the future obstacle distance increases in every prediction step.
 
\section{Evaluation}
\label{sec:evaluation}
We apply our obstacle avoidance approach to the UAV described in \cite{schleich2021icuas}.
It was specially designed for search and rescue missions and is based on the DJI Matrice 210 v2 platform equipped with an Ouster OS-0 3D-LiDAR.
The experiments were done using the DJI Hardware-in-the-Loop simulation and Gazebo \cite{koenig2004iros}.
The velocity commands generated by our method were executed using the onboard DJI flight controller.

\begin{table}
\caption{ Results using random velocity commands for the indoor environment (\reffig{fig:indoor_arena}). We report the average flight velocity $v_\text{avg}$, as well as the minimal and average obstacle distances $d_\text{min}$ and $d_\text{avg}$. 
As an indicator for the proneness to local minima, we additionally report the average distances to the commanded waypoints $d_\text{target}$ at which the UAV stopps.}
\begin{center}
\begin{tabular}[t]{l|ccccc}
\multirow{2}{*}{Method}  & $d_\text{target}$ & $v_\text{avg}$ & $d_\text{min}$ & $d_\text{avg}$ \\
  & \lbrack\si{\meter}\rbrack & \lbrack\si{\meter\per\second}\rbrack & \lbrack\si{\meter}\rbrack & \lbrack\si{\meter}\rbrack \\  \hline
 PF \cite{beul2018ral} & -  & \textcolor{red}{\bf{2.38}} & 0.00 & 1.62 \\
 TG \cite{beul2020ssrr} & \bf{27.05} & \bf{1.70} & 1.21 & 2.30 \\
 Ours ($r_\text{vel}$) & 27.11 & 1.02 & 1.30 & 2.05 \\
 Ours ($r_\text{euclidean}$) & 30.80 & 1.27 & 1.20 & 2.40 \\
 Ours ($r_\text{vel}$+pred.)& 27.73 & 1.56 & \bf{1.41} & \bf{2.59}
\end{tabular}
\end{center}
 \vspace{-0.3cm}
\label{tab:random}
\end{table}

We evaluate our approach against the potential field method (PF) from \cite{beul2018ral}, where two influence spheres around obstacles are used.
In the larger one, \ie the passive avoidance sphere, repulsive forces slow down the UAV movement into the obstacle direction, while the smaller active avoidance sphere is used to push the UAV away from the obstacle.
Additionally, we compare against the trajectory generation method (TG) from \cite{beul2020ssrr}, which generates time-optimal trajectories to sampled target points until a collision-free trajectory is found.
To evaluate our different design decisions, we additionally apply two variants of our method to all experiments.
First, we use the Euclidean distance $r_\text{euclidean}$ instead of the velocity-dependent future range value $r_\text{vel}$ for the support definition of angular repulsive forces.
Second, we remove our trajectory prediction and estimate the time-to-contact based on the vector projection of the commanded velocity onto the obstacle directions.

For all experiments, we use the following parameters:
\begin{center}
\begin{tabular}[t]{c|c|c|c|c}
$d_\text{safe}$ & $d_\text{close}$ & $a_\text{max}$ & $t_\text{contact}$ & $d_\text{min\_contact}$\\ \hline
\SI{1.5}{\meter} & \SI{1.0}{\meter} & \SI{2}{\meter\per\second} &  \SI{1.5}{\second}  &  \num{2}
\end{tabular} .
\end{center}
Our method starts decelerating the UAV if the predicted obstacle distance falls below the safety threshold $d_\text{safe}$ within the look-ahead $t_\text{contact}$.
Thus, the radius of the passive avoidance sphere for the potential field method \cite{beul2018ral} is chosen as $d_\text{passive} = d_\text{safe} + v_\text{max} t_\text{contact}$, where $v_\text{max}$ denotes the maximum allowed flight velocity.
The size of the active avoidance sphere corresponds to our safety threshold, \ie $d_\text{active} = d_\text{safe}$.

\begin{table}
\newcommand{\mc}[3]{\multicolumn{#1}{#2}{#3}}
\caption{ Results for the path following experiment (\cf\reffig{fig:path_following} and \reffig{fig:support_trajectories}). Success denotes whether the path was fully tracked $\checkmark$, the method got stuck in a local minimum $(\checkmark)$ or a collision occurred $\times$. Additionally, we report the length of the flight path $l_\text{path}$ (until the UAV stops, either due to reaching the end of the target path, getting stuck in a local minima or due to a collision), the average flight velocity $v_\text{avg}$, as well as the minimal and average obstacle distances $d_\text{min}$ and $d_\text{avg}$. }
\begin{center}
\begin{tabular}[t]{c|l|c|cccc}
$v_\text{max}$ &  \multirow{2}{*}{Method} &  \multirow{2}{*}{Suc.} & $l_\text{path}$ & $v_\text{avg}$ & $d_\text{min}$ & $d_\text{avg}$ \\  
 \lbrack\si{\meter\per\second}\rbrack &  &  & \lbrack\si{\meter}\rbrack & \lbrack\si{\meter\per\second}\rbrack & \lbrack\si{\meter}\rbrack & \lbrack\si{\meter}\rbrack \\  \hline
 \multirow{6}{*}{1} & PF \cite{beul2018ral} & $(\checkmark)$ & 81.27 & 0.91 & 1.04 & 1.84 \\
 & TG \cite{beul2020ssrr} & $(\checkmark)$ & 102.00 & 0.87 & 1.24 & 1.76 \\
 & Ours ($r_\text{vel}$) & $(\checkmark)$ & 125.17 & 0.77 & 1.37 & 1.79 \\ 
 & Ours ($r_\text{euclidean}$) & $\checkmark$ & 125.36 & 0.91 & 1.37 & 1.72 \\ 
 & Ours ($r_\text{vel}$+pred.) & $\checkmark$ & \bf{124.89} & \bf{0.95} & \bf{1.38} & \bf{1.91} \\ \hline
 
 \multirow{6}{*}{2} & PF \cite{beul2018ral} & $\checkmark$ & 130.19 & 1.85 & 0.46 & 1.75 \\
 & TG \cite{beul2020ssrr} & $(\checkmark)$ & 104.13 & 1.71 & 1.31 & 1.79 \\
 & Ours ($r_\text{vel}$) & $(\checkmark)$ & 126.37 & 0.92 & 1.37 & 1.77 \\ 
 & Ours ($r_\text{euclidean}$) & $\checkmark$ & 128.51 & \bf{1.97} & \bf{1.40} & \bf{2.12} \\ 
 & Ours ($r_\text{vel}$+pred.) & $\checkmark$ & \bf{127.88} & 1.89 & 1.39 & 2.01 \\ \hline
 
 \multirow{6}{*}{3} & PF \cite{beul2018ral} & $\checkmark$ & 133.83 & \bf{2.79} & 0.31 & 1.72 \\
 & TG \cite{beul2020ssrr} & $(\checkmark)$ & 106.89 & 2.21 & 1.20 & 1.80 \\
 & Ours ($r_\text{vel}$) & $(\checkmark)$ & 125.02 & 0.84 & 1.37 & 1.75 \\ 
 & Ours ($r_\text{euclidean}$) & $\checkmark$ & \bf{133.82} & 2.14 & 1.35 & 1.83 \\ 
 & Ours ($r_\text{vel}$+pred.) & $\checkmark$ & 134.48 & 2.77 & \bf{1.39} & \bf{2.43} \\ \hline
 
 \multirow{6}{*}{4} & PF \cite{beul2018ral} & $\times$ & 123.30 & \textcolor{red}{\bf{3.70}} & 0.00 & 1.63 \\
 & TG \cite{beul2020ssrr} & $(\checkmark)$ & 3.67 & 1.29 & 1.29 & 1.81 \\
 & Ours ($r_\text{vel}$) & $(\checkmark)$ & 126.07 & 0.87 & 1.37 & 1.76  \\ 
 & Ours ($r_\text{euclidean}$) & $\checkmark$ & \bf{134.04} & 2.50 & 1.34 & 1.93  \\ 
 & Ours ($r_\text{vel}$+pred.) & $\checkmark$ & 140.00 & \bf{3.48} & \bf{1.39} & \bf{2.55} \\ \hline
 
 \multirow{6}{*}{5} & PF \cite{beul2018ral} & $\times$ & 73.82 & \textcolor{red}{\bf{4.57}} & 0.00 & 1.61 \\
 & TG \cite{beul2020ssrr} & $(\checkmark)$ & 50.44 & 1.60 & 1.36 & 1.74 \\
 & Ours ($r_\text{vel}$) & $(\checkmark)$ & 126.12 & 0.88 & 1.37 & 1.77 \\ 
 & Ours ($r_\text{euclidean}$) & $(\checkmark)$  & 56.91 & 2.08 & \bf{1.45} & 1.75 \\ 
 & Ours ($r_\text{vel}$+pred.) & $\checkmark$ & \bf{142.33} & \bf{3.84} & 1.33 & \bf{2.45} \\ \hline
 
 \multirow{6}{*}{6} & PF \cite{beul2018ral} & $\times$ & 72.89 & \textcolor{red}{\bf{4.90}} & 0.00 & 1.58\\
 & TG \cite{beul2020ssrr} & $(\checkmark)$ & 135.06 & 2.59 & 1.38 & 1.79 \\
 & Ours ($r_\text{vel}$) & $(\checkmark)$  & 103.93 & 0.85 & 1.37 & 1.75 \\ 
 & Ours ($r_\text{euclidean}$) & $(\checkmark)$ & 57.44 & 1.97 & \bf{1.44} & 1.88 \\ 
 & Ours ($r_\text{vel}$+pred.) & $\checkmark$ & \bf{144.75} & \bf{4.29} & 1.39 & \bf{2.42} \\ \hline
\end{tabular}
\end{center}
\label{tab:path_following}
\vspace{-0.3cm}
\end{table}

\begin{figure*}
 \begin{center}
    \includegraphics[width=0.8\linewidth]{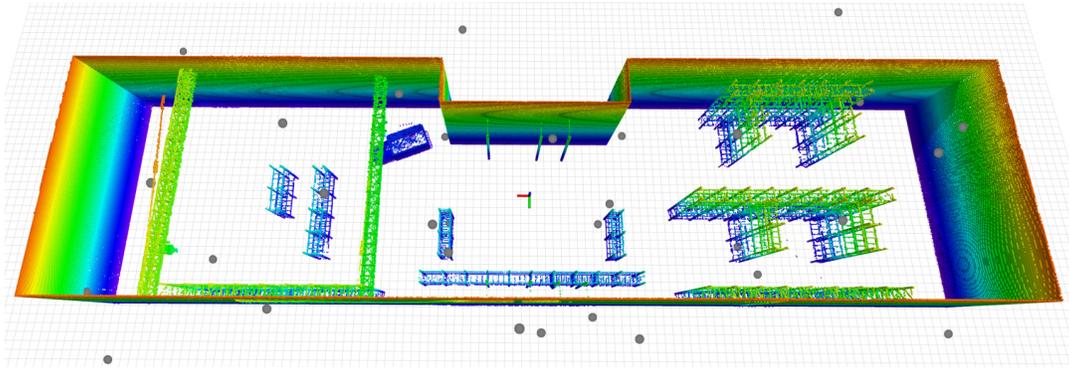}
 \end{center}
 \caption{A simulated indoor environment for the random input experiment. The randomly chosen waypoints are marked by gray spheres. The initial UAV position is at the map center, marked by the coordinate axes.}
 \label{fig:indoor_arena}
\end{figure*}

Obstacle avoidance methods must ensure a safe distance to obstacles independent of the---possibly adversarial---input commands.
In a first experiment, we thus investigate how the different methods react to randomly chosen velocity commands.
For this, we designed the indoor warehouse environment shown in \reffig{fig:indoor_arena}.
The UAV starts at the center of the map with an altitude of \SI{2}{\meter} and is commanded to move at a velocity of \SI{3}{\meter\per\second} into the direction of a randomly sampled target position, which might even be outside the warehouse.
Every \SI{10}{\second}, a new target is sampled.

\reftab{tab:random} summarizes the results for a flight time of \SI{300}{\second} per method. 
The flight velocities were too high for the potential field method \cite{beul2018ral} and it collided with an obstacle.
While all other methods successfully avoided collisions, our method maintained the largest distance to obstacles.
Using the velocity-dependent range value $r_\text{vel}$ for the repulsive force definition significantly increases the obstacle distance and the trajectory prediction allows higher velocities.

\begin{figure*}
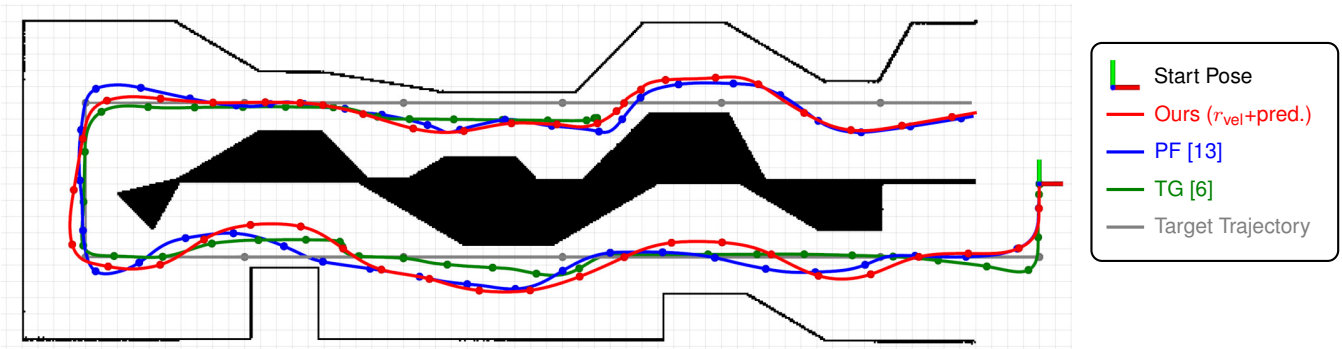

\begin{center}
\begin{tikzpicture}[
 	font=\sffamily\footnotesize,
    every node/.append style={text depth=.2ex},
	l/.style={font=\sffamily\tiny},
]

\def\imageWidth{14.7};
\def\imageHeight{4.0};
\def\margin{0.2};

\definecolor{ours}{rgb}     {1.0,0.0,0.0}
\definecolor{pf}{rgb}   {0.0,0.0,1.0}
\definecolor{tg}{rgb}    {0.0,0.5,0.0}
\definecolor{target}{rgb}    {0.5,0.5,0.5}

\node[anchor=south west,inner sep=0] (image_range) at (0,0) {\includegraphics[width=0.8\textwidth]{img/vel_3.png}};

\draw[thick, rounded corners] (\imageWidth - \margin,\imageHeight+\margin) -- (\textwidth,\imageHeight+\margin) -- (\textwidth,6/16*\imageHeight-\margin) -- (\imageWidth-\margin,6/16*\imageHeight-\margin) -- cycle;
\node[anchor=north west,inner sep=0] (legend) at (\imageWidth,\imageHeight) {\includegraphics[width=0.025\textwidth]{figures/axes.png}};
\node[right, shift={(0.28,0.)}] at (legend) {Start Pose};
\draw[very thick, ours] (\imageWidth,13/16*\imageHeight) -- (\imageWidth+0.5,13/16*\imageHeight) node[right, shift={(0.0,0.)}] {Ours ($r_\text{vel}$+pred.)};
\draw[very thick, pf] (\imageWidth,11/16*\imageHeight) -- (\imageWidth+0.5,11/16*\imageHeight) node[right, shift={(0.0,0.)}] {PF \cite{beul2018ral}};
\draw[very thick, tg] (\imageWidth,9/16*\imageHeight) -- (\imageWidth+0.5,9/16*\imageHeight) node[right, shift={(0.0,0.)}] {TG \cite{beul2020ssrr} };
\draw[very thick, target] (\imageWidth,7/16*\imageHeight) -- (\imageWidth+0.5,7/16*\imageHeight) node[right, shift={(0.0,0.)}] {Target Trajectory};

\end{tikzpicture}
 \end{center}
 \caption{Top-down view of example trajectories for our method (red), the potential field method from \cite{beul2018ral} (blue) and the trajectory generation method \cite{beul2020ssrr} (green). On each trajectory, spheres are sampled with a time difference of \SI{1}{\second}. The target trajectory (gray) shall be tracked at a velocity of \SI{3}{\meter\per\second}. Our method shows the most far-sighted behavior and thus also maintains the largest distances to obstacles.}
 \label{fig:path_following}
 \vspace{-0.2cm}
\end{figure*}

\begin{figure*}
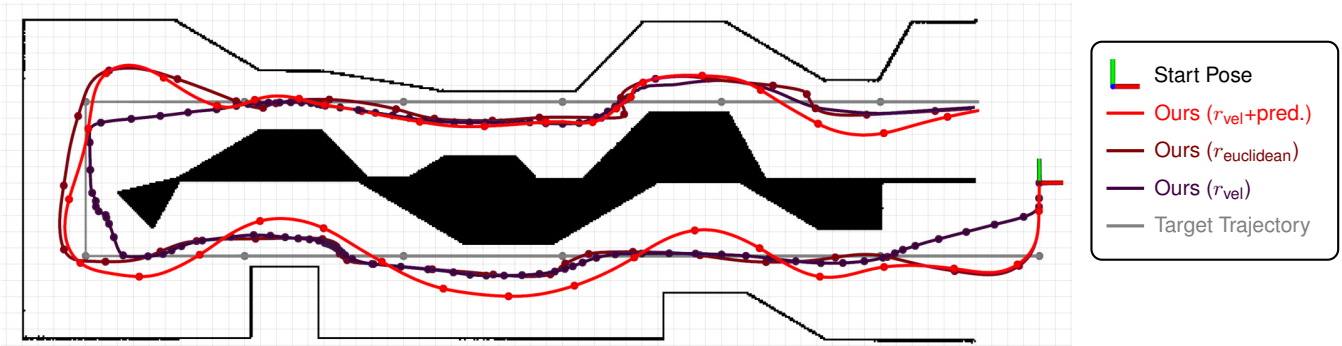

\begin{center}
\begin{tikzpicture}[
 	font=\sffamily\footnotesize,
    every node/.append style={text depth=.2ex},
	l/.style={font=\sffamily\tiny},
]

\def\imageWidth{14.7};
\def\imageHeight{4.0};
\def\margin{0.2};

\definecolor{ours}{rgb}     {1.0,0.0,0.0}
\definecolor{euclidean}{rgb}   {0.5,0.0,0.0}
\definecolor{vel}{rgb}    {0.25,0.0,0.25}
\definecolor{target}{rgb}    {0.5,0.5,0.5}

\node[anchor=south west,inner sep=0] (image_range) at (0,0) {\includegraphics[width=0.8\textwidth]{img/support_vel_4.png}};

\draw[thick, rounded corners] (\imageWidth - \margin,\imageHeight+\margin) -- (\textwidth,\imageHeight+\margin) -- (\textwidth,6/16*\imageHeight-\margin) -- (\imageWidth-\margin,6/16*\imageHeight-\margin) -- cycle;
\node[anchor=north west,inner sep=0] (legend) at (\imageWidth,\imageHeight) {\includegraphics[width=0.025\textwidth]{figures/axes.png}};
\node[right, shift={(0.28,0.)}] at (legend) {Start Pose};
\draw[very thick, ours] (\imageWidth,13/16*\imageHeight) -- (\imageWidth+0.5,13/16*\imageHeight) node[right, shift={(0.0,0.)}] {Ours ($r_\text{vel}$+pred.)};
\draw[very thick, euclidean] (\imageWidth,11/16*\imageHeight) -- (\imageWidth+0.5,11/16*\imageHeight) node[right, shift={(0.0,0.)}] {Ours ($r_\text{euclidean}$)};
\draw[very thick, vel] (\imageWidth,9/16*\imageHeight) -- (\imageWidth+0.5,9/16*\imageHeight) node[right, shift={(0.0,0.)}] {Ours ($r_\text{vel}$)};
\draw[very thick, target] (\imageWidth,7/16*\imageHeight) -- (\imageWidth+0.5,7/16*\imageHeight) node[right, shift={(0.0,0.)}] {Target Trajectory};

\end{tikzpicture}
 \end{center}
 \caption{Top-down view of example trajectories for our method (bright red), our method without velocity-dependent support radius (dark red), and our method without trajectory prediction (purple). On each trajectory, spheres are sampled with a time difference of \SI{1}{\second}. The target trajectory (gray) shall be tracked at a velocity is \SI{4}{\meter\per\second}. Combining the velocity-dependent support radius with trajectory prediction results in a smooth trajectory at high velocities. }
 \label{fig:support_trajectories}
\vspace{-0.2cm}
\end{figure*}

In a second experiment, we analyze at which velocities the different obstacle avoidance methods can track a pre-planned path that is too close to (initially unknown) obstacles.
The results are summarized in \reftab{tab:path_following} and \reffig{fig:path_following} depicts example trajectories. 
Our method shows a more far-sighted behavior than the Cartesian potential field method and the sampling-based trajectory generation.
Thus, our method can maintain larger distances to the obstacles, even at high flight velocities.
Furthermore, it is less prone to local minima.
\reffig{fig:support_trajectories} compares trajectories from the different variants of our method.
The velocity-dependent support radius results in smoother trajectories, while the trajectory prediction allows significantly higher flight velocities.

\begin{figure*}
\begin{center}
\begin{tikzpicture}[
 	font=\sffamily\footnotesize,
    every node/.append style={text depth=.2ex},
	l/.style={font=\sffamily\tiny},
]

\def\imageWidth{14.7};
\def\imageHeight{1.6};
\def\margin{0.2};

\definecolor{ours}{rgb}     {1.0,0.0,0.0}
\definecolor{pf}{rgb}   {0.0,0.0,1.0}
\definecolor{tg}{rgb}    {0.0,0.5,0.0}
\definecolor{target}{rgb}    {0.5,0.5,0.5}

\node[anchor=south west,inner sep=0] (image_range) at (0,0) {\includegraphics[width=0.8\textwidth]{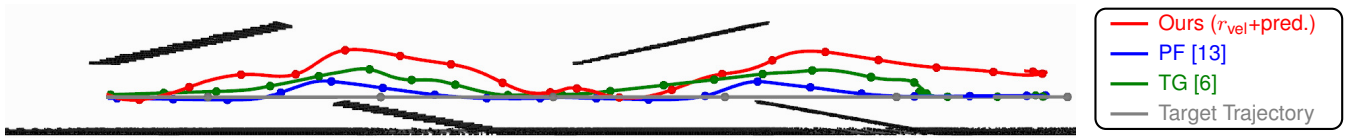}};

\draw[thick, rounded corners] (\imageWidth - \margin,\imageHeight+\margin) -- (\textwidth,\imageHeight+\margin) -- (\textwidth,1/4*\imageHeight-\margin) -- (\imageWidth-\margin,1/4*\imageHeight-\margin) -- cycle;
\draw[very thick, ours] (\imageWidth,\imageHeight) -- (\imageWidth+0.5,\imageHeight) node[right, shift={(0.0,0.)}] {Ours ($r_\text{vel}$+pred.)};
\draw[very thick, pf] (\imageWidth,3/4*\imageHeight) -- (\imageWidth+0.5,3/4*\imageHeight) node[right, shift={(0.0,0.)}] {PF \cite{beul2018ral}};
\draw[very thick, tg] (\imageWidth,2/4*\imageHeight) -- (\imageWidth+0.5,2/4*\imageHeight) node[right, shift={(0.0,0.)}] {TG \cite{beul2020ssrr} };
\draw[very thick, target] (\imageWidth,1/4*\imageHeight) -- (\imageWidth+0.5,1/4*\imageHeight) node[right, shift={(0.0,0.)}] {Target Trajectory};

\end{tikzpicture}
 \end{center}
 \caption{ Side view of an excerpt of the environment with vertical obstacles. Shown are the trajectories of our method (red), the potential field method from \cite{beul2018ral} (blue) and the trajectory generation method \cite{beul2020ssrr} (green). On each trajectory, spheres are sampled with a time difference of \SI{1}{\second}. The target trajectory (from right to left) is depicted in gray. The maximal allowed velocity is \SI{3}{\meter\per\second}.}
 \label{fig:slopes}
\end{figure*}

\begin{table}
\caption{ Results for the path following experiment with vertical obstacles (\cf\reffig{fig:slopes}). We report the length of the flight path $l_\text{path}$, the average flight velocity $v_\text{avg}$, the flight duration $t_\text{flight}$, as well as the minimal and average obstacle distances $d_\text{min}$ and $d_\text{avg}$. }
\begin{center}
\begin{tabular}{l|ccccc}
\multirow{2}{*}{Method}  & $l_\text{path}$ & $v_\text{avg}$ & $t_\text{flight}$ & $d_\text{min}$ & $d_\text{avg}$ \\  
 & \lbrack\si{\meter}\rbrack & \lbrack\si{\meter\per\second}\rbrack & \lbrack\si{\second}\rbrack & \lbrack\si{\meter}\rbrack & \lbrack\si{\meter}\rbrack \\  \hline
PF \cite{beul2018ral}  & 128.13 & 2.85 & 44.96 & 1.03 & 1.68 \\
TG \cite{beul2020ssrr} & 130.63 & 2.61 & 50.05 & 1.57 & 1.97 \\
Ours ($r_\text{vel}$) &  \bf{124.98} & 1.53 & 81.67 & \bf{1.59} & 2.16  \\ 
Ours ($r_\text{euclidean}$) &  129.20 & \bf{3.00} & \bf{43.07} & 1.47 & 1.94 \\ 
Ours ($r_\text{vel}$+pred.) & 130.20 & 2.92 & 44.59 & 1.29 & \bf{2.59} \\
\end{tabular}
\end{center}
\label{tab:path_following_slopes}
\end{table}

One advantage of our proposed angular potential field-based obstacle avoidance method is the possibility to adapt the flight direction to nearby obstacles without affecting the velocity.
To further evaluate this aspect, we designed an experiment where the UAV moves through a corridor with changing height values for both, floor and ceiling.
Thus, the vertical flight angle has to be frequently adjusted to keep a safe distance to obstacles.
\reftab{tab:path_following_slopes} shows that the trajectory prediction is essential for our method to obtain high flight velocities which are about twice the velocities of our method without prediction.
Interestingly, the classic potential field method achieves similar velocities in this experiment.
This is due to the fact that it does not adjust its trajectory to the obstacles much, as can be seen in \reffig{fig:slopes}.
Our angular potential field method, however, maximizes the obstacle distance while still achieving fast flights.

\begin{table}
\caption{ Comparison of the computations times per control iteration. }
\begin{center}
\begin{tabular}{l|ccc}
\multirow{2}{*}{Method}  & $t_\text{min}$ & $t_\text{avg}$ & $t_\text{max}$ \\  
 & \lbrack\si{\milli\second}\rbrack & \lbrack\si{\milli\second}\rbrack & \lbrack\si{\milli\second}\rbrack \\  \hline
PF \cite{beul2018ral}  & $\mathbf{<1}$ & \bf{4} & \bf{11} \\
TG \cite{beul2020ssrr} & 15 & 24 & 227 \\ 
Ours & 1 & 15 & 32 \\
\end{tabular}
\end{center}
\vspace{-0.4cm}
\label{tab:computation_times}
\end{table}

Finally, we compare the computation times of each method in \reftab{tab:computation_times}.
As expected, the classic potential field method is the fastest one.
The sampling-based trajectory generation method shows a large runtime variance between different control iterations.
Especially in constraint environments, it can be difficult to sample valid target points, resulting in high maximal computation times.
Our method reliably achieves runtimes significantly below the sensor measurement frequency of \SI{20}{\hertz}.

\section{Conclusion}
\label{sec:conclusion}
In this work, we proposed a fast predictive obstacle avoidance method that combines angular potential fields to determine the flight direction with trajectory prediction and time-to-contact estimation to scale the flight velocity magnitude.
Our method operates directly on 3D LiDAR range images and does not depend on higher-level localization or mapping.
The conducted experiments show that the proposed method reliably keeps a safe distance to obstacles, while showing a more far-sighted behavior than previous obstacle  avoidance methods that directly operate on sensor data.
Thus, our method is suitable for high-velocity flights in cluttered environments.

\bibliographystyle{IEEEtranBST/IEEEtran}
\bibliography{literature}

\begin{thebibliography}{10}
\providecommand{\url}[1]{#1}
\csname url@rmstyle\endcsname
\providecommand{\newblock}{\relax}
\providecommand{\bibinfo}[2]{#2}
\providecommand\BIBentrySTDinterwordspacing{\spaceskip=0pt\relax}
\providecommand\BIBentryALTinterwordstretchfactor{4}
\providecommand\BIBentryALTinterwordspacing{\spaceskip=\fontdimen2\font plus
\BIBentryALTinterwordstretchfactor\fontdimen3\font minus
  \fontdimen4\font\relax}
\providecommand\BIBforeignlanguage[2]{{%
\expandafter\ifx\csname l@#1\endcsname\relax
\typeout{** WARNING: IEEEtran.bst: No hyphenation pattern has been}%
\typeout{** loaded for the language `#1'. Using the pattern for}%
\typeout{** the default language instead.}%
\else
\language=\csname l@#1\endcsname
\fi
#2}}

\bibitem{schleich2021icuas}
D.~Schleich, M.~Beul, J.~Quenzel, and S.~Behnke, ``Autonomous flight in unknown
  {GNSS}-denied environments for disaster examination,'' in \emph{International
  Conference on Unmanned Aircraft Systems (ICUAS)}, 2021.

\bibitem{oleynikova2016iros}
H.~Oleynikova, M.~Burri, Z.~Taylor, J.~Nieto, R.~Siegwart, and E.~Galceran,
  ``Continuous-time trajectory optimization for online {UAV} replanning,'' in
  \emph{IEEE/RSJ International Conference on Intelligent Robots and Systems
  (IROS)}, 2016.

\bibitem{usenko2017iros}
V.~Usenko, L.~Von~Stumberg, A.~Pangercic, and D.~Cremers, ``Real-time
  trajectory replanning for {MAVs} using uniform {B}-splines and a {3D}
  circular buffer,'' in \emph{IEEE/RSJ International Conference on Intelligent
  Robots and Systems (IROS)}, 2017.

\bibitem{zhang2018iros}
J.~Zhang, R.~G. Chadha, V.~Velivela, and S.~Singh, ``{P-CAP: P}re-computed
  alternative paths to enable aggressive aerial maneuvers in cluttered
  environments,'' in \emph{IEEE/RSJ International Conference on Intelligent
  Robots and Systems (IROS)}, 2018.

\bibitem{zhang2019iros}
J.~Zhang, C.~Hu, R.~G. Chadha, and S.~Singh, ``Maximum likelihood path planning
  for fast aerial maneuvers and collision avoidance,'' in \emph{IEEE/RSJ
  International Conference on Intelligent Robots and Systems (IROS)}, 2019.

\bibitem{beul2020ssrr}
M.~Beul and S.~Behnke, ``Trajectory generation with fast lidar-based {3D}
  collision avoidance for agile {MAVs},'' in \emph{IEEE International Symposium
  on Safety, Security and Rescue Robotics (SSRR)}, 2020.

\bibitem{fox1997ram}
D.~Fox, W.~Burgard, and S.~Thrun, ``The dynamic window approach to collision
  avoidance,'' \emph{IEEE Robotics \& Automation Magazine}, vol.~4, no.~1, pp.
  23--33, 1997.

\bibitem{missura2019icra}
M.~Missura and M.~Bennewitz, ``Predictive collision avoidance for the dynamic
  window approach,'' in \emph{IEEE International Conference on Robotics and
  Automation (ICRA)}, 2019.

\bibitem{dobrevski2020iros}
M.~Dobrevski and D.~Sko{\v{c}}aj, ``Adaptive dynamic window approach for local
  navigation,'' in \emph{IEEE/RSJ International Conference on Intelligent
  Robots and Systems (IROS)}, 2020.

\bibitem{missura2022iros}
M.~Missura, A.~Roychoudhury, and M.~Bennewitz, ``Fast-replanning motion control
  for non-holonomic vehicles with aborted {A*},'' in \emph{IEEE/RSJ
  International Conference on Intelligent Robots and Systems (IROS)}, 2022.

\bibitem{khatib1986ijrr}
O.~Khatib, ``Real-time obstacle avoidance for manipulators and mobile robots,''
  in \emph{The International Journal of Robotics Research (IJRR)}, vol.~5,
  no.~1, 1986, pp. 90--98.

\bibitem{nieuwenhuisen2013predictive}
M.~Nieuwenhuisen, M.~Schadler, and S.~Behnke, ``Predictive potential
  field-based collision avoidance for multicopters,'' \emph{Int. Arch.
  Photogramm. Remote Sens. Spatial Inf. Sci}, vol.~1, p.~W2, 2013.

\bibitem{beul2018ral}
M.~Beul, D.~Droeschel, M.~Nieuwenhuisen, J.~Quenzel, S.~Houben, and S.~Behnke,
  ``Fast autonomous flight in warehouses for inventory applications,''
  \emph{IEEE Robotics and Automation Letters (RA-L)}, vol.~3, no.~4, pp.
  3121--3128, 2018.

\bibitem{sezer2012ras}
V.~Sezer and M.~Gokasan, ``A novel obstacle avoidance algorithm:“{F}ollow the
  gap method”,'' \emph{Robotics and Autonomous Systems}, vol.~60, no.~9, pp.
  1123--1134, 2012.

\bibitem{houshyari2021robotica}
H.~Houshyari and V.~Sezer, ``A new gap-based obstacle avoidance approach:
  follow the obstacle circle method,'' \emph{Robotica}, pp. 1--24, 2021.

\bibitem{cho2018jat}
J.-H. Cho, D.-S. Pae, M.-T. Lim, and T.-K. Kang, ``A real-time obstacle
  avoidance method for autonomous vehicles using an obstacle-dependent gaussian
  potential field,'' \emph{Journal of Advanced Transportation}, vol. 2018.

\bibitem{koenig2004iros}
N.~Koenig and A.~Howard, ``Design and use paradigms for {G}azebo, an
  open-source multi-robot simulator,'' in \emph{IEEE/RSJ International
  Conference on Intelligent Robots and Systems (IROS)}, 2004.

\end{thebibliography}

\end{document}